\documentclass[runningheads]{llncs}

\usepackage{algorithmic}
\usepackage{graphicx}
\usepackage{textcomp}
\usepackage{xcolor}
\usepackage{subcaption}
\usepackage{tabularx}
\usepackage{soul}
\usepackage{diagbox}

\begin{document}

%%
%% Title and short title for page headers
\title{A Methodology for Graphical Comparison of Regression Models}
\titlerunning{A Methodology for Graphical Comparison of Regression Models}

\author{Nassime Mountasir\inst{1,2} \and
Baptiste Lafabregue\inst{1} \and
Bruno Albert\inst{2} \and
Nicolas Lachiche\inst{1}}

\authorrunning{N. Mountasir et al.} % Version courte pour les en-têtes de page

\institute{Laboratoire ICube, Université de Strasbourg, 300 Bd Sébastien Brant, 67400 Illkirch-Graffenstaden, France \and
ENGLAB, Technology \& Strategy Engineering, 3 rue Denis Papin, 67300 Schiltigheim, France\\
\email{nassime.mountasir@icube.unistra.fr}}

\maketitle

%%
%% Abstract
\begin{abstract}
To properly assess the performance of regression models, it is crucial to use effective evaluation methods. In regression, performance is usually measured using various metrics such as MAE or RMSE, that provide numerical summaries of predictive accuracy. However, while these metrics are widely used in the literature for summarizing model performance and useful to distinguish between models performing poorly and well, they often aggregate information too much.
We address these limitations by introducing a novel visualization approach that highlights key aspects of regression model performance, illustrated on three real datasets to assess its relevance. The proposed method is built in two steps: selecting the best models with 1D visualizations, and considering the errors of the chosen models in a 2D space. The so-called 2D Error Space is made upon three components: 1) the geometry of the space itself, 2) the use of a colormap to visualize the percentile-based distribution of errors, which facilitates the identification of dense regions and outliers and 3) the application of the Mahalanobis distance to account for correlations and differences in scale for comparison. By graphically representing the distribution of errors, this provides a more comprehensive view of models' performance, enabling users to uncover patterns that traditional aggregating metrics may obscure.

\keywords{visualization, regression, comparison, metrics, analysis}
\end{abstract}

\section{Introduction}

Regression is one of the most fundamental and extensively studied problems in machine learning. Over time, the landscape of regression modeling has evolved considerably — from simple linear models and kernel-based approaches to increasingly complex architectures such as ensemble methods and deep neural networks.
Each new generation of models improves predictive capability but also increases complexity, making configuration, interpretability, and fair comparison more challenging. Although performance is typically summarized by scalar metrics such as MSE, RMSE, MAE, or $R^2$ \cite{an}, these measures offer only a limited view of a model’s actual behavior.
The focus of this article is on evaluating the predictions and qualifying the errors of one or more models for a regression problem, irrespective of the nature of the problem or the input data. The primary concern is the comparison between the model predictions and the actual observed values, and especially the limitations of established evaluation methods. To illustrate this, we use several datasets trained on multiple models, each providing actual values along with predictions. The complete implementation used to train the models and their definition is available here: \url{https://anonymous.4open.science/r/visualization-tools}.

\section{Limits of usual metrics and visualizations}

Standard metrics efficiently filter underperforming models but often mask critical behavioral differences between competitive candidates. We identify three limitations where aggregate scores conceal essential details, thus failing to differentiate models with distinct error distributions.

\subsection{Metrics}

Let us consider the performance of 12 regression models as measured by MAE and RMSE on a Dataset A as reported in Table~\ref{tab:mae_rmse_values}.
As a reminder, the error for each prediction of a model is calculated as the difference $r$ between the predicted value  $\hat{y}$ and the actual value $y$, as: $r=\hat{y}-y$. This error can be positive (an overestimation) when $\hat{y} > y$, or negative (underestimation) when $ \hat{y} < y$.

Some models exhibit excellent results, with both low MAE and RMSE, eg. models A1, A2, A9 or A10,
while others perform poorly with both metrics significantly higher, such as models A3 or A5.
As models A1, A2, A9 and A10 have metrics that are significantly lower than the other models, we could say that those models tend to overperform compared to the others, leaving only four models to compare instead of twelve.

\begin{table}[ht]
    \caption{MAE and RMSE values for Dataset A}
    \label{tab:mae_rmse_values}
    \centering
    \footnotesize
    \begin{tabular}{|l|l|l|l|l|l|l|l|l|l|l|l|l|l|}
    \hline
    \diagbox{\textbf{Metric}}{\textbf{Model}} & \textbf{A1} & \textbf{A2} & \textbf{A3} & \textbf{A4} & \textbf{A5} & \textbf{A6} & \textbf{A7} & \textbf{A8} & \textbf{A9} & \textbf{A10} & \textbf{A11} & \textbf{A12} \\ \hline
    \textbf{MAE} & 12.9 & 13.5 & 32.3 & 21.3 & 35.6 & 21.8 & 18.1 & 19.3 & 10.9 & 11.0 & 20.9 & 18.2 \\ \hline
    \textbf{RMSE} & 16.8 & 17.5 & 37.2 & 25.7 & 40.7 & 26.3 & 22.7 & 24.8 & 14.7 & 14.5 & 24.8 & 23.2 \\ \hline
    \end{tabular}
\end{table}

However, while these metrics are relevant for summarizing model performance and for discriminating between very good and very bad models, it becomes difficult to use them when their values are close. In particular, they aggregate too much information, potentially obscuring important details about the nature and distribution of prediction errors. 
For example, in critical applications such as medical diagnosis or autonomous driving, it is crucial to avoid extreme values to prevent potentially life-threatening decisions, whereas for financial forecasting or marketing strategies, more stable models might be preferred even if they occasionally produce outliers, as long as their overall predictive performance remains high.
We identify three limitations in the use of metrics to compare two models.

\begin{figure}[ht]
    \begin{subfigure}[b]{0.3\textwidth}
        \includegraphics[width=\textwidth]{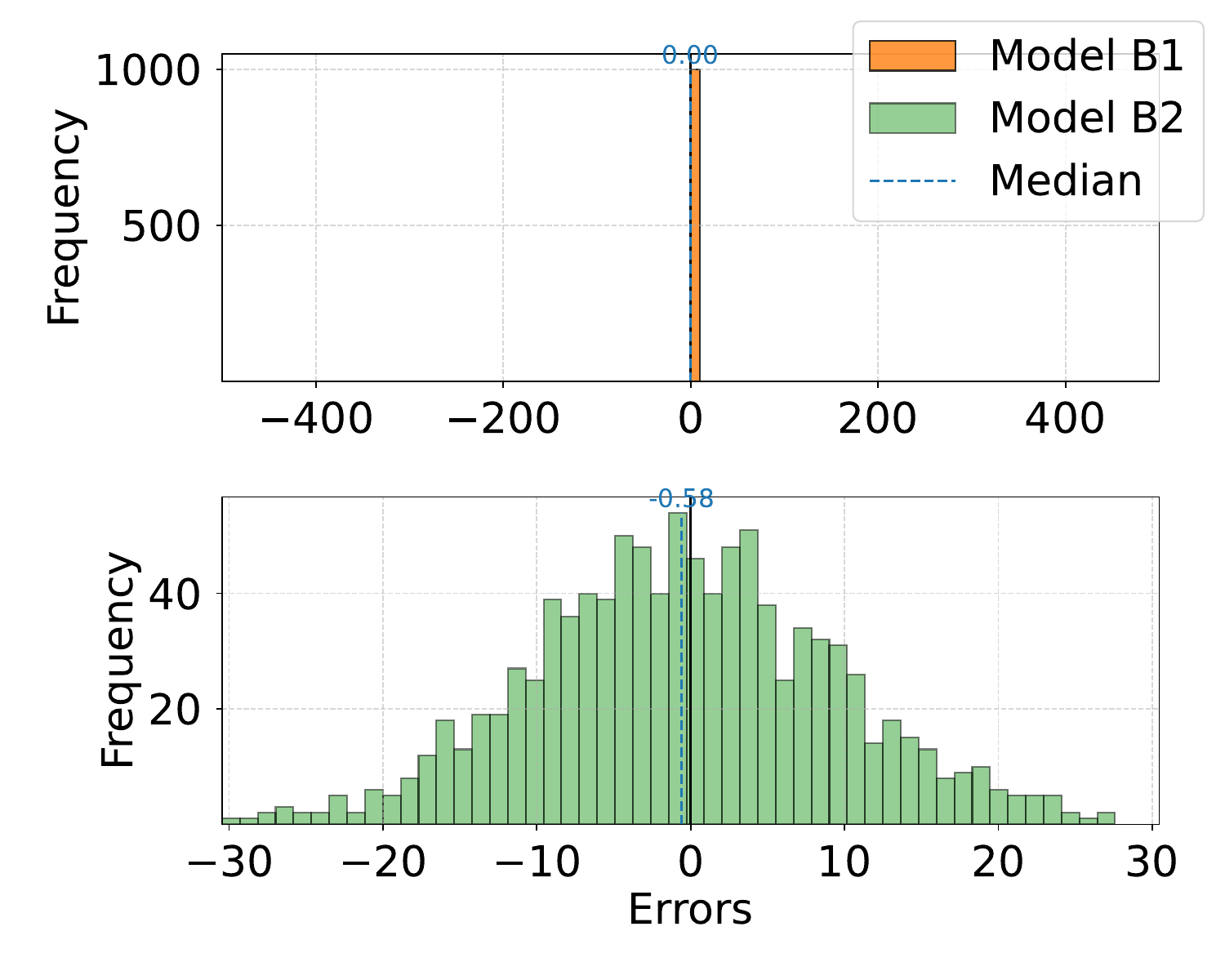}
        \caption{Extreme outlier vs. Moderate errors}
        \label{fig:distribution_test}
    \end{subfigure}
    \hfill
    \begin{subfigure}[b]{0.3\textwidth}
        \includegraphics[width=\textwidth]{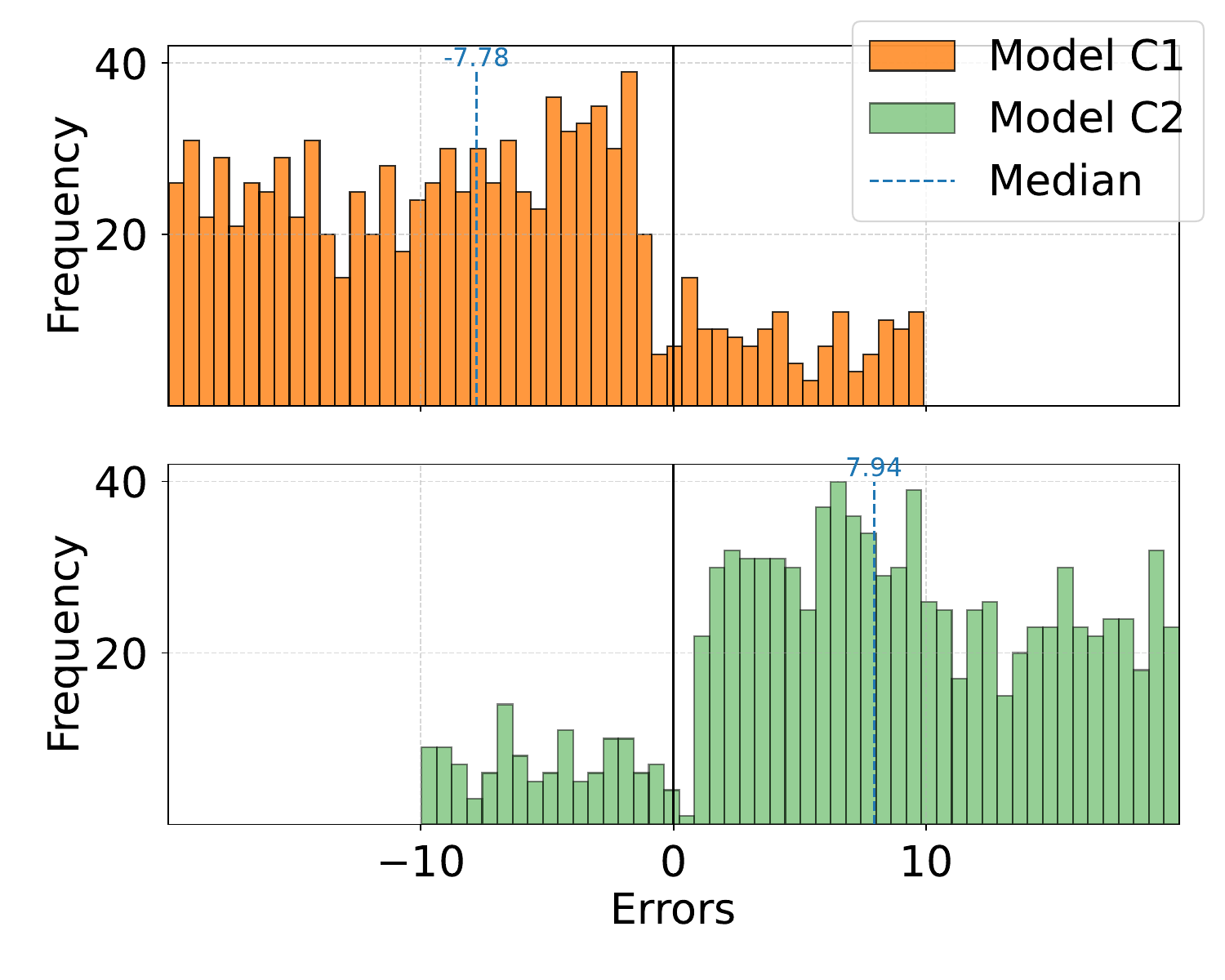}
        \caption{Underestimation vs. overestimation patterns}
        \label{fig:two_distributions_test}
    \end{subfigure}
    \hfill
    \begin{subfigure}[b]{0.3\textwidth}
        \includegraphics[width=\textwidth]{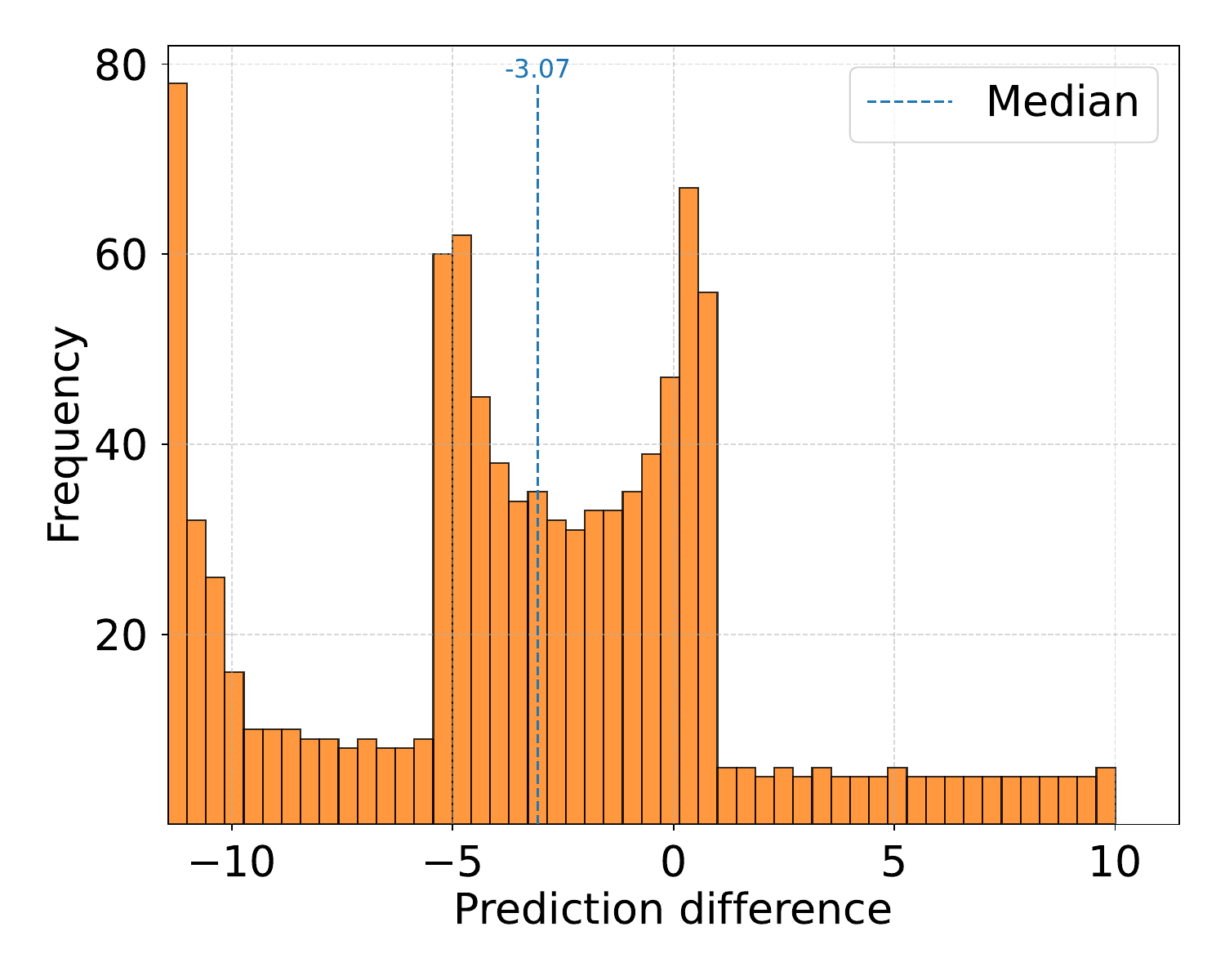}
        \caption{Prediction differences between two models} 
        \label{fig:distribution_diff}
    \end{subfigure}
    \hfill
    \caption{Overview of the three main limitations of aggregate error metrics}
\end{figure}

\paragraph{Moderate vs. Extreme errors}

Consider a Dataset B and 2 models B1 and B2 with distributions shown in Figure~\ref{fig:distribution_test}. For those models, B1 has a MAE of 0.5 and a RMSE of 15.81, where B2 has a MAE of 7.86 with a RMSE of 9.88.
Here, MAE favors B1 while RMSE favors B2, showing that the perceived performance of a model can change depending on how strongly outliers are taken into account.

\paragraph{Under and over estimations}

Standard metrics rely on absolute or squared errors, which inherently masks the directionality of predictions and makes it impossible to distinguish between under- and over-estimation. This limitation is clearly illustrated in Figure~\ref{fig:two_distributions_test}, where Model C1 (consistently underestimating) and Model C2 (consistently overestimating) produce nearly identical MAE (9.02 vs. 9.34) and RMSE (10.57 vs. 10.94) values, despite their completely opposite error distributions

\paragraph{Similar errors on different individuals}

Aggregate metrics can conceal cases where models behave differently despite similar MAE or RMSE values. Comparing predictions directly through the deviation $\delta(m_1, m_2) = \hat{y}_1 - \hat{y}_2$ helps reveal these differences. Figure~\ref{fig:distribution_diff} shows that although Models D1 and D2 yield comparable MAE (3.18 vs. 3.19) and RMSE (3.76 vs. 4.39) values, their deviation histogram exposes clear discrepancies: D1 consistently predicts lower values, while D2’s underestimations are more dispersed. Since MAE and RMSE ignore directional and instance-level differences, visualization techniques are necessary to reveal error patterns that aggregated metrics cannot capture, motivating the method introduced in this article.

\subsection{Visualizations}
\begin{figure}[ht]
    \begin{subfigure}[b]{0.5\textwidth}
        \includegraphics[width=0.8\textwidth]{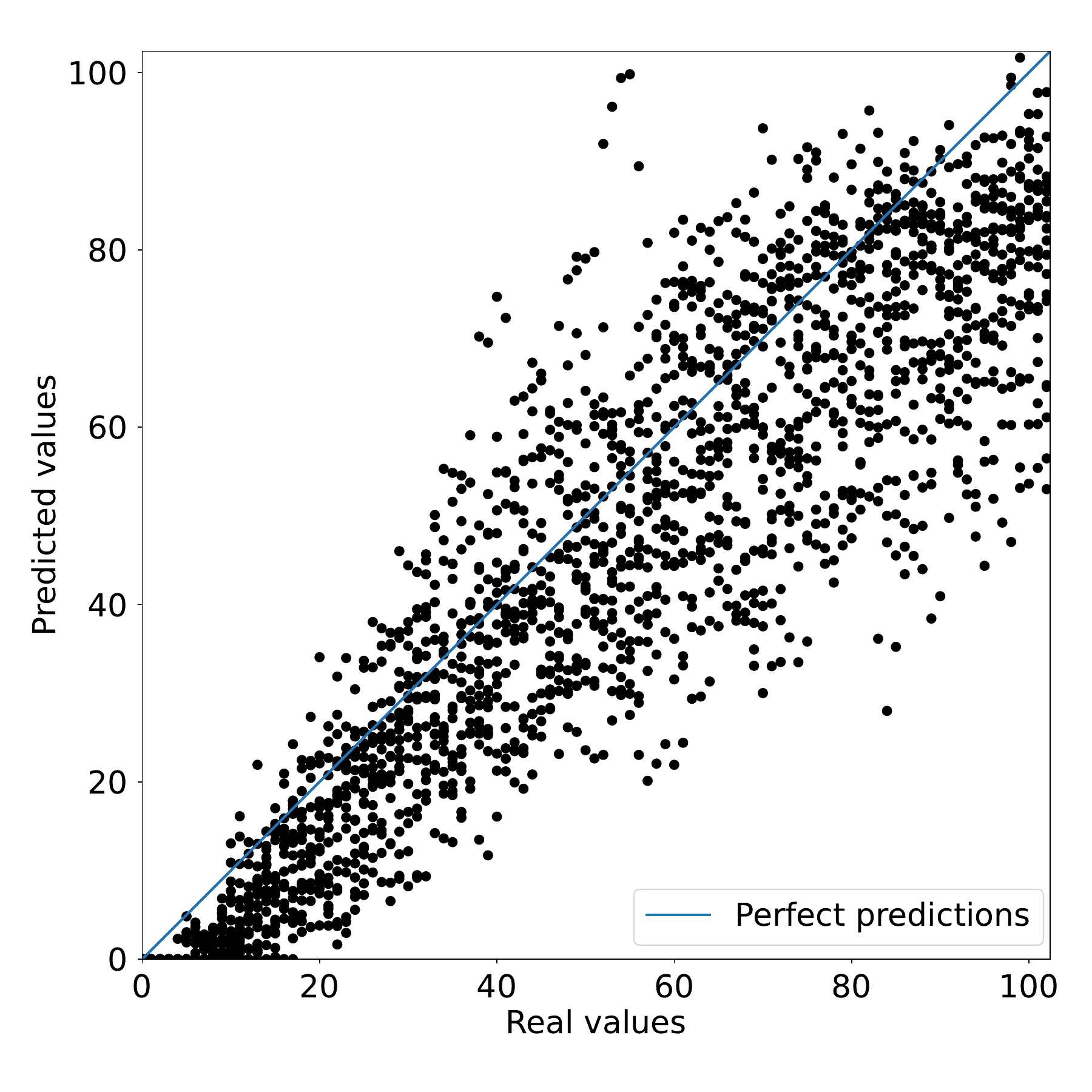}
        \caption{Model A1}
        \label{fig:actual_predicted_a}
    \end{subfigure}
    \hfill
    \begin{subfigure}[b]{0.5\textwidth}
        \includegraphics[width=0.8\textwidth]{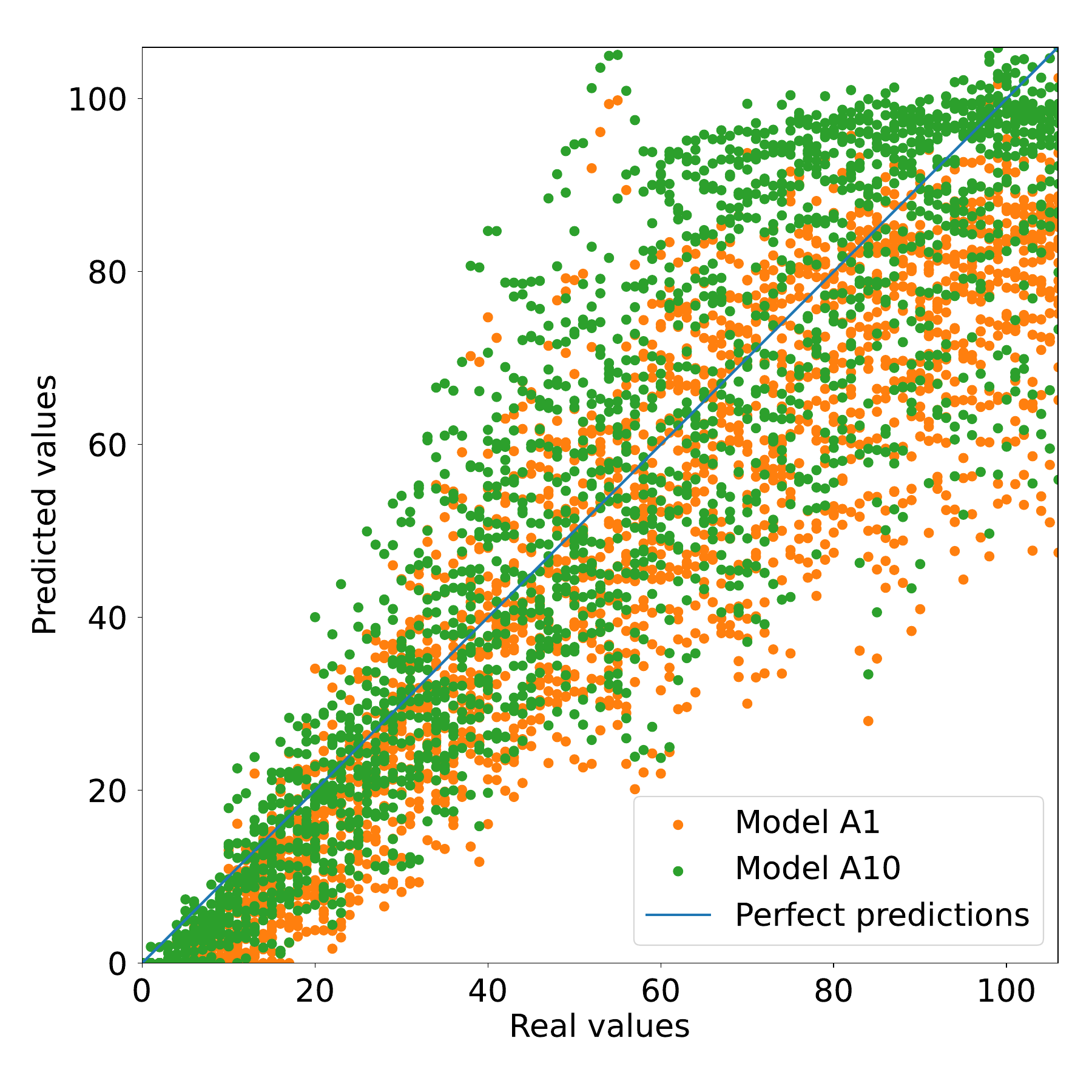}
        \caption{Model A1 and A10}
        \label{fig:actual_predicted_b}
    \end{subfigure}
    \hfill
    \caption{Comparison of the predictions and the real values for Dataset A}
    \label{fig:actual_predicted}
\end{figure}

%Bl : faudrait référencer les sub figures, je t'ai ajouté les labels (\ref{fig:actual_predicted_a})
%NM : c'est fait
As mentioned before, a common way to compare regression models graphically is by displaying predicted values alongside real values in a scatter plot.
The Figure~\ref{fig:actual_predicted_a} shows the comparison between predictions and real values for a model.
One approach to compare two models would be to plot the predictions of two models as a function of the ground truth, to see if their ranges differ, as done in Figure~\ref{fig:actual_predicted_b}. 
In this figure, both distributions are plotted, with in orange the errors of the first model and in green the errors of the second one.
However, using only predictions does not provide an accurate overview of the models' performance, as it is entirely possible to have similar models that make significantly different mistakes.
Additionally, it shares the same drawback as the Figure~\ref{fig:actual_predicted_a}, especially as we don't have any information about the number of points.
To address these limitations, we propose a visual method that compares models by plotting error distributions against each other.

\section{State of the art}

To contextualize our approach, we first review the current state of the art in regression model evaluation, focusing on widely used performance metrics and visualization techniques used to enhance the comprehension of model predictions and errors.
Metrics remain the primary indicators of model performance~\cite{japkowicz2024machine}, and recent studies such as \cite{ma2024dgcd,ye2024uncertainty} continue to rely on them to compare regression models.
However, the visualization of regression model performance has evolved significantly, incorporating a variety of techniques to enhance the understanding of predictive accuracy and error distributions. 

Visualization techniques have long been used to complement numerical summaries. Scatter plots are commonly used to compare predicted and actual values, which are known as residual plots. However, they tend to become difficult to interpret with larger datasets due to point overlap and density issues. For such cases, hexagonal binning (hexbin plots) is a common technique to control overplotting: points are aggregated into hexagonal cells and colored by count or an aggregate statistic, improving the perception of density and structure in such scatter data.
However, hexbins still fail to capture how points are structured relative to a central reference or distribution shape as shown in Section~\ref{section:proximity}.
Histograms and density plots offer a way to visualize the distribution of prediction errors, making it easier to identify skewness and outliers~\cite{eda}.
Other simple summary graphics, such as box-and-whisker plots, commonly known as boxplots, are widely used to compare the distribution of errors across models as they succinctly display medians, interquartile ranges and extreme values.
Specialized visual summaries have been proposed for multi-criteria evaluation. 
Taylor diagrams summarize multiple aspects of model performance, such as correlation, standard deviation, and RMSE, providing a comprehensive overview in a single diagram~\cite{taylor}.
Overall, the main issue is that traditional visualizations show results of one model for individual data points rather than providing a clear comparison between two models across the dataset, limiting their usefulness for assessing relative performance.

\section{Graphical comparison of the errors of regression models}

To compare regression models effectively, we propose a two-step visual analysis of their prediction errors. The goal is first to quickly identify underperforming models (Section~\ref{section:1d}) and then study the most promising ones in more detail (Section~\ref{section:2d}).

\subsection{1D Comparison}
\label{section:1d}

A straightforward way to compare models with one-dimensional tools is to use boxplots that summarize the range and distribution of prediction errors for each model, facilitating the comparison of performance.

\begin{figure}[h]
    \centering
    \begin{subfigure}[t]{0.35\textwidth}
        \centering
        \includegraphics[width=\textwidth]{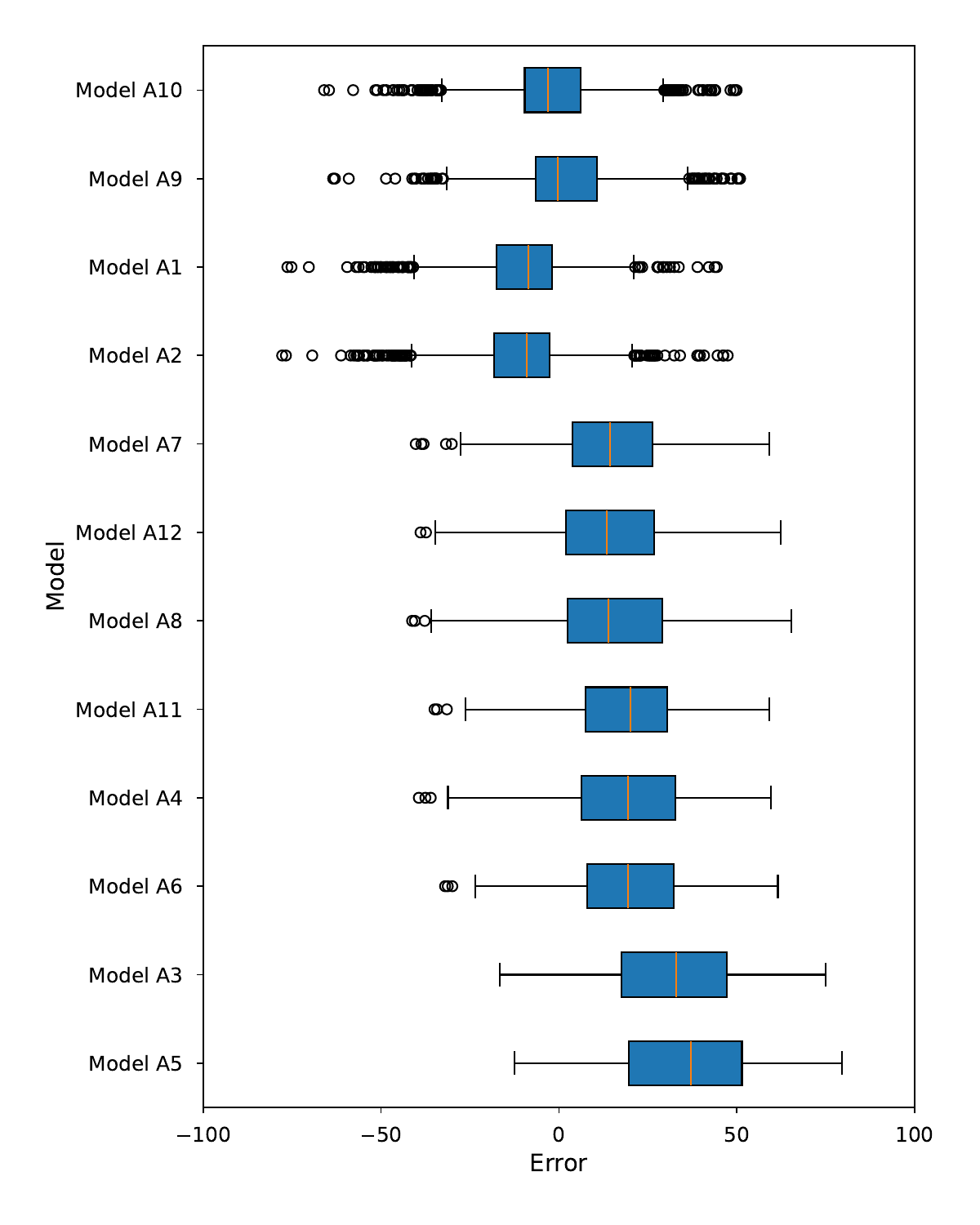}
        \caption{Prediction error boxplots}
        \label{fig:boxplots}
    \end{subfigure}
    \hfill
    \begin{subfigure}[t]{0.56\textwidth}
        \centering
        \includegraphics[width=\textwidth]{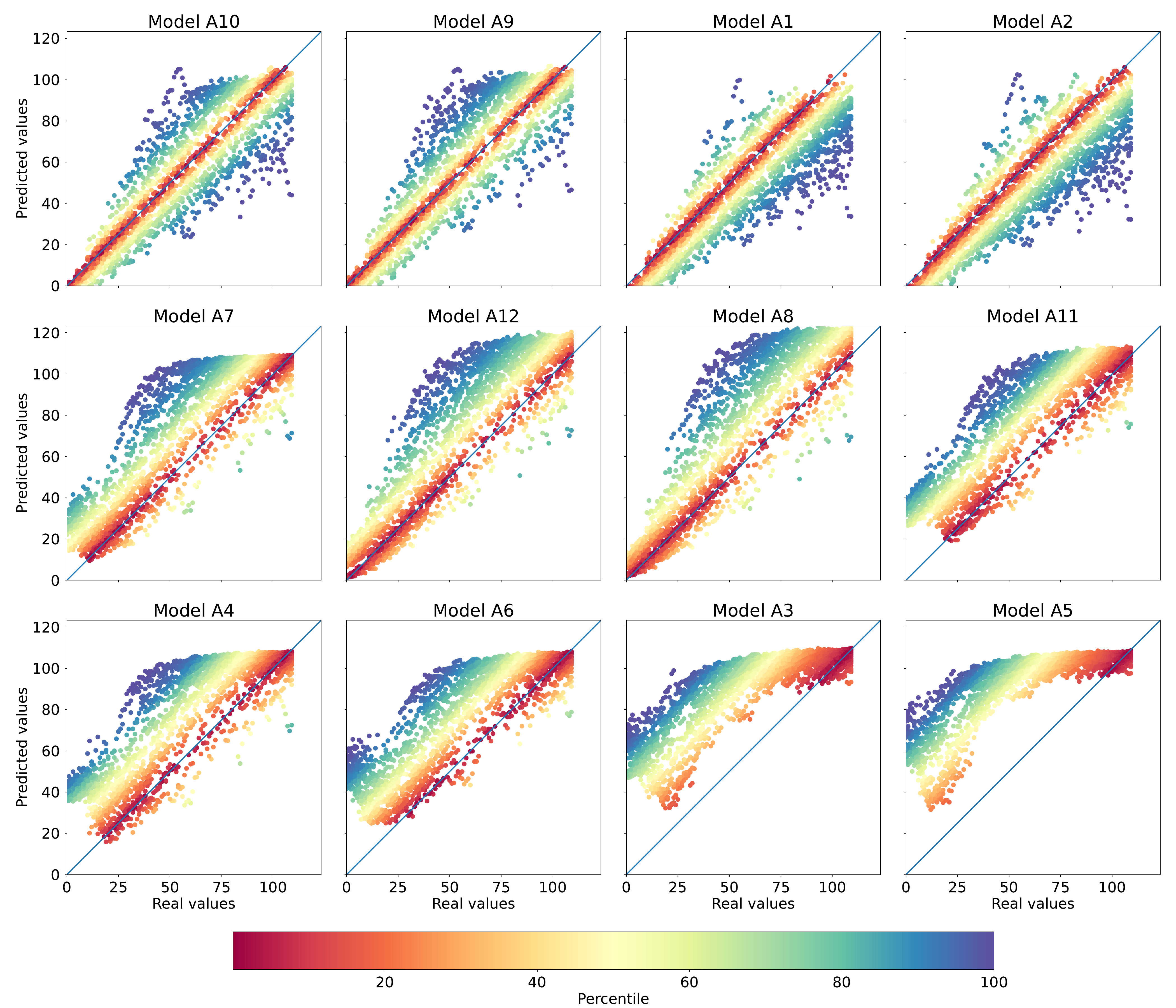}
        \caption{Predicted vs.\ real values for all models}
        \label{fig:predicted_real_grid}
    \end{subfigure}

    \caption{Comparison of error distributions (left) and predicted vs. real values (right).}
    \label{fig:combined_boxplot_scatter}
\end{figure}

In Figure~\ref{fig:boxplots}, we display the errors of all models on Dataset A. We can choose to sort them by a certain metric to visualize better the performance depending on a chosen criterion; here, the models are sorted by RMSE to maximize the influence of outliers and highlight those with the most stable performance.

Each boxplot shows the spread of errors: the narrower the box, the more consistent the predictions are. Outliers are also visible, helping to identify models that occasionally make large errors.
We can see that models A11, A10, A1 and A2 generally predict well, as shown with the metrics, but they have differences on the way they predict. Model A1 and A2 sometimes predict too low values, whereas models A9 and A10 are more likely to predict higher values compared to them.
It can also allow us to identify patterns that are not visible with metrics only. For instance, even though Model A8 is supposedly a decent model, it tends to make large errors on some individuals, in a manner similar to models A6 or A7.
To go further and better understand how the models behave across the full range of predictions, we can look at scatter plots of predicted values versus true values. Since it is difficult to compare multiple models on the same plot, we generate one plot per model.
Figure~\ref{fig:predicted_real_grid} shows the scatter plots for all 12 models trained on Dataset A.
To make the errors easier to see, we use a color scale: warm colors indicate accurate predictions, and cool colors indicate large errors. This helps identify areas where models perform well and where they struggle.

We can see that the models that are performing best have yellow outlines around the diagonal, as most points are close to it. "Best" models - i.e the first rows in the Figure - generally have closer points to the diagonal and less points very far from it, while "less-performing" models have the propensity to have a lot of points that are outliers. Models A1 and A2 are for instance very similar, whereas models A9 and A10 have a different pattern with both underestimations and overestimations, as points are located above and under the diagonal.
On the other hand, models A3, A5 and A6 have very few points close to the diagonal, and Model A7 has a lot of outliers, showing that their performance is not very good.
Finally, Models A4, A11 and A12 have a more nuanced profile of errors with less stretched values around the diagonal but some outliers.
This visualization gives more context than metrics alone. It shows, for example, if a model is accurate on low values but less reliable on higher ones, or if it tends to make specific types of errors. These patterns are difficult to detect using MAE or RMSE alone.
With such visualizations, we could say that Model A1 and Model A10 are the ones performing well but behave differently depending on the individuals, highlighting the need for a more detailed comparison.

\subsection{2D Error Space}
\label{section:2d}

\begin{figure}[htbp]
    \centering
    \includegraphics[width=0.3\columnwidth]{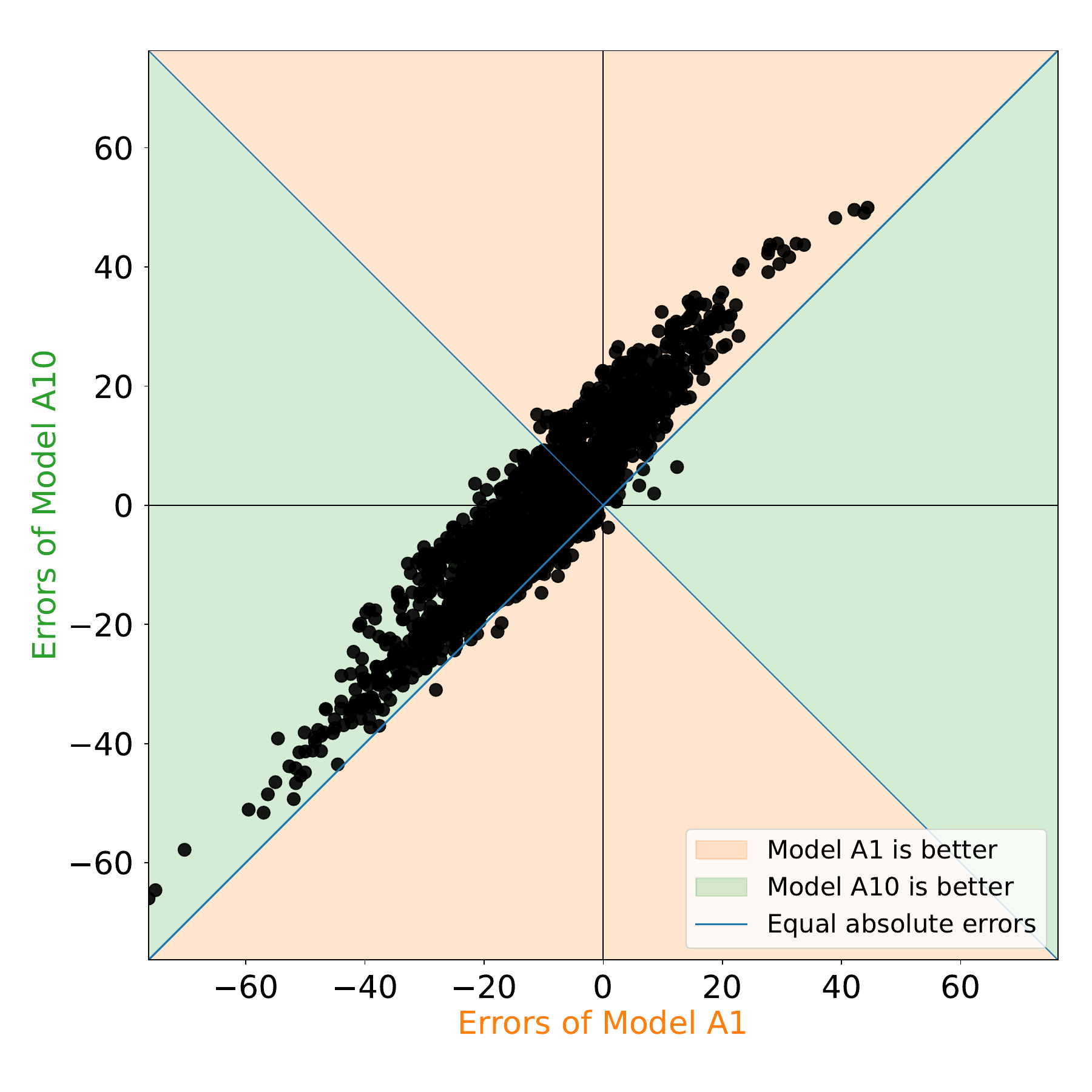}
    \caption{Comparison of the errors of Model A1 and Model A10}
    \label{fig:2d_space}
\end{figure}

To compare two models, we propose to visualize the errors of one model with respect to the other in a plot, called the 2D Error Space.
For this section, we choose Models A1 and A10 as they are performing both well and show a very different profile of errors.
Figure~\ref{fig:2d_space} shows the representation of this space for the two models.
Each point corresponds to the errors of predictions for a given individual, where the x-axis is the error of one model and the y-axis represents the error of the other model.
In this space, the diagonal $y = x$ represents the points where the error of the models is the same. For situations where one of the models is overestimating as much as
the other is underestimating, we can also use the diagonal $y = -x$.

The diagonals represent equal absolute errors for the models, so we can define two regions: one where the first model is better, \textit{i.e.} its absolute error is smaller, and another where the second model is better, creating two hourglasses. These regions are respectively colored in orange and green. We will refer to these spaces as "comparison zones" throughout the rest of the article. These hourglasses allow us to effectively identify regions where there are more points than another, in case one contains significantly more points than another.
The difference between the models for an individual can be visualized by looking at the closest diagonal from the corresponding point.
Finally, the horizontal and vertical axes provide information about whether the first and second model tend to underestimate or overestimate.
The points located to the left and to the right of the vertical axis are the points where Model A1 is respectively underestimating and overestimating, and the points located below and above the horizontal axis 
are the points where the Model A10 is respectively underestimating and overestimating.
Such a visualization allows us to directly assess the performance of the models relative to each other, as we have both the information on the predictions made and the difference from the ground truth. This provides insight into error distribution and whether the models tend to make large errors
or predictions close to reality.

\subsection{Density and proximity with median}
\label{section:proximity}

Using a scatter plot to visualize error distribution in the comparison zones can make it difficult to clearly identify the areas where most points are concentrated.
This difficulty is especially pronounced with large datasets, where points may overlap. To solve this issue, we can represent the density of the points using a colormap.

\begin{figure}[htbp]
    \begin{subfigure}[b]{0.3\textwidth}
        \includegraphics[width=\textwidth]{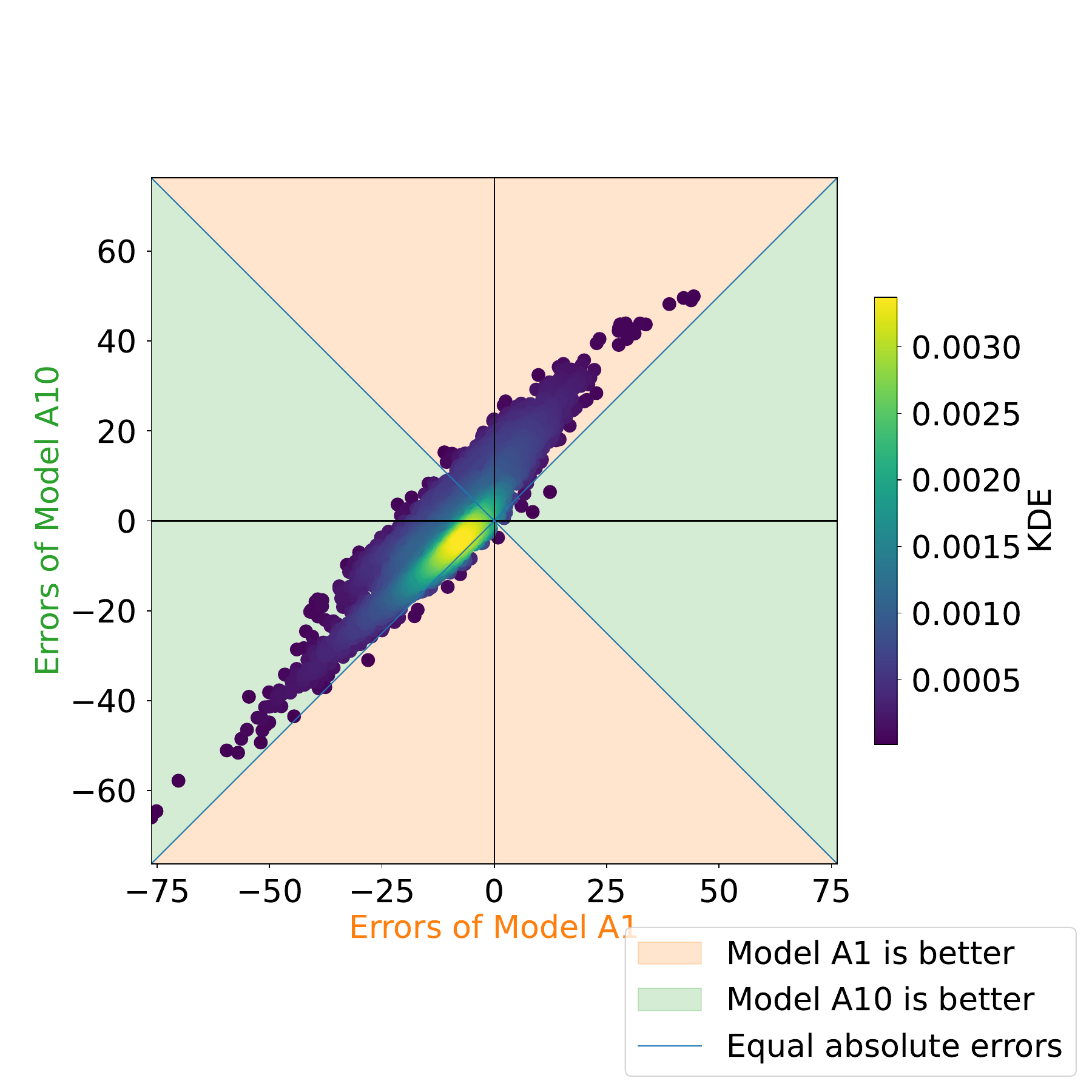}
        \caption{KDE}
        \label{fig:kde}
    \end{subfigure}
    \hfill
    \begin{subfigure}[b]{0.3\textwidth}
        \includegraphics[width=\textwidth]{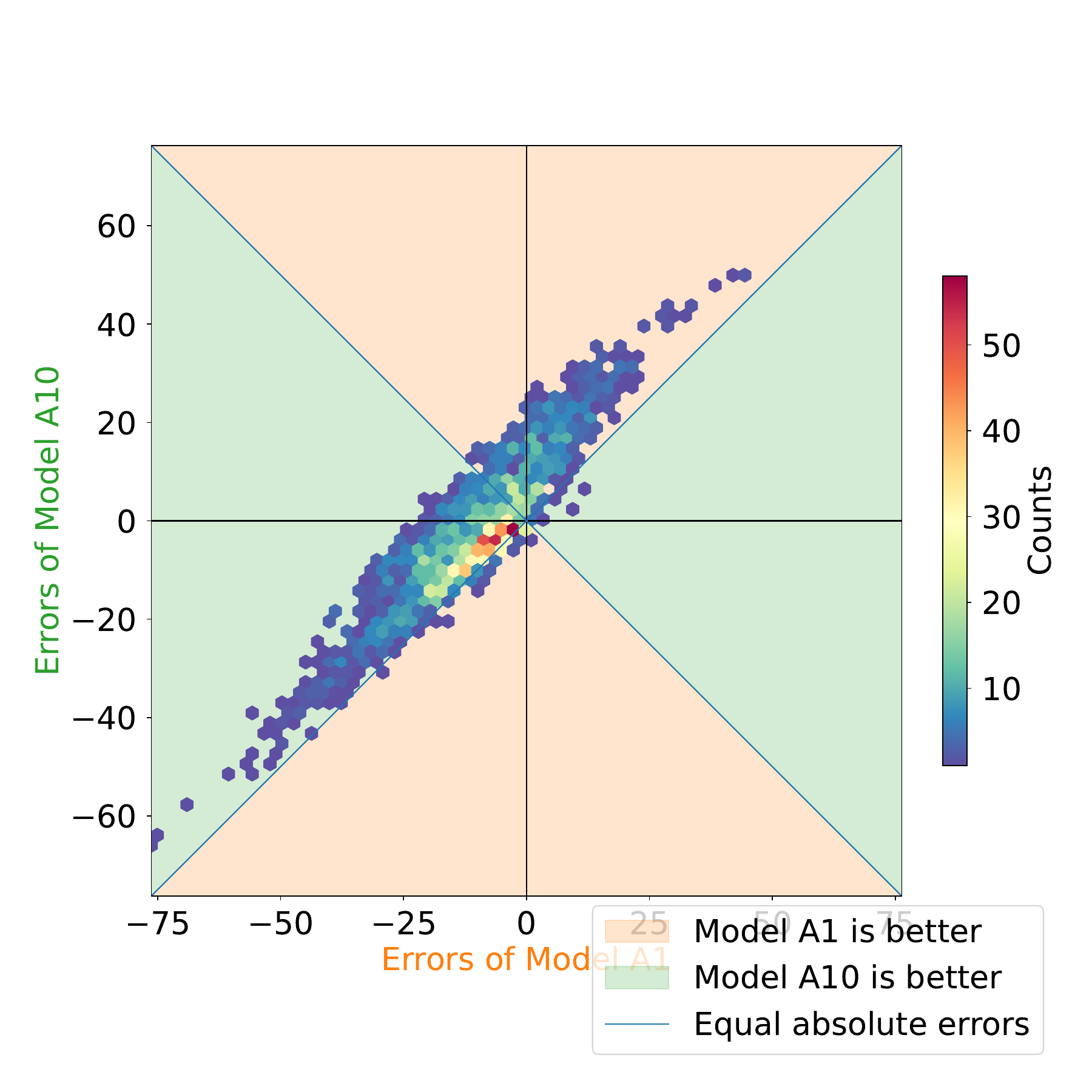}
        \caption{Hexbin}
        \label{fig:hexbin_example}
    \end{subfigure}
    \hfill
    \begin{subfigure}[b]{0.3\textwidth}
        \includegraphics[width=\textwidth]{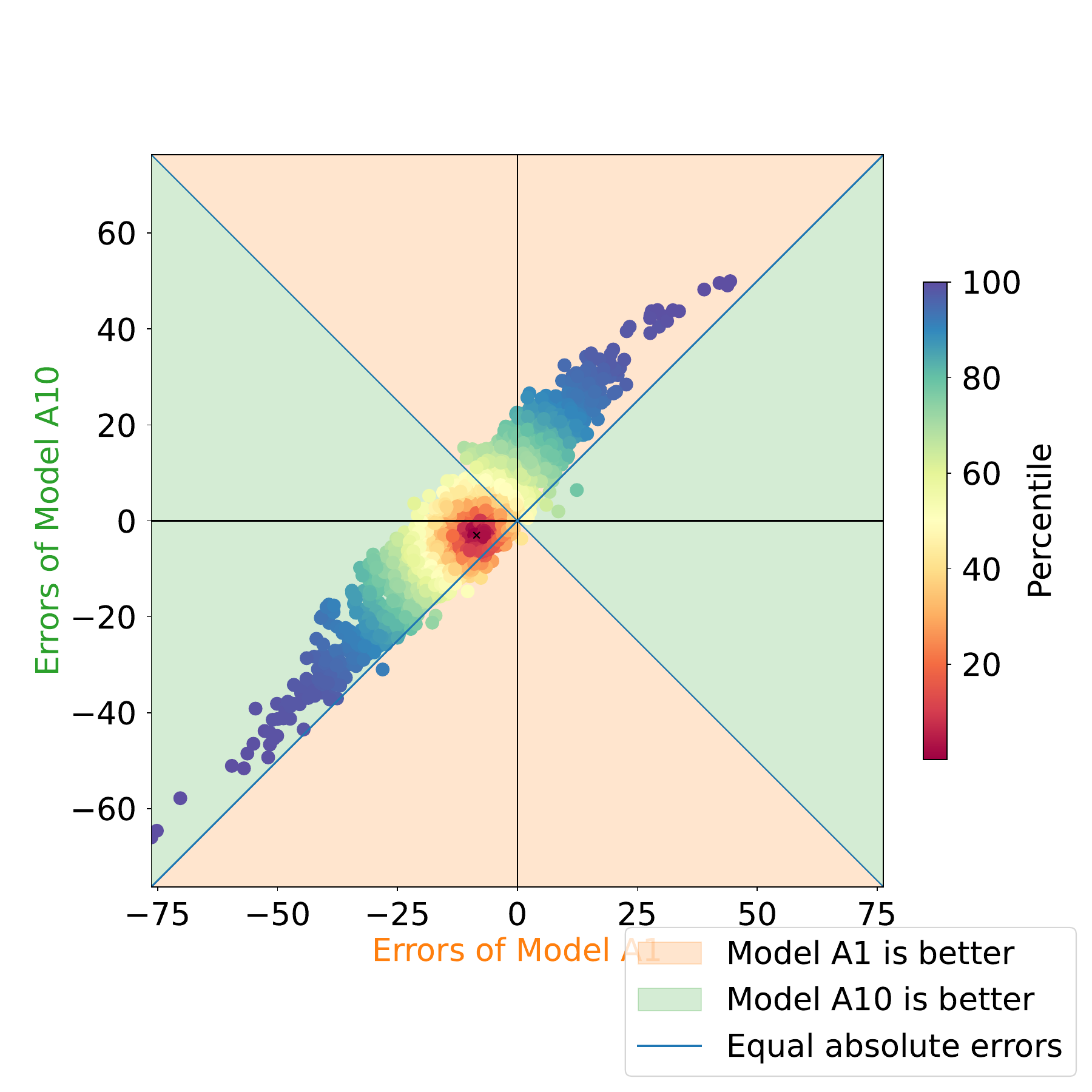}
        \caption{Proximity}
        \label{fig:proximity}
    \end{subfigure}
    \hfill
    \caption{Representation of the proximity of points with different methods}
\end{figure}

The colormap in Figure~\ref{fig:kde} illustrates the density of points, calculated by Kernel Density Estimation (KDE)~\cite{chen2017tutorial}.
However, while KDE-based visualizations provide a useful estimation of point density, they also present several limitations.
KDE produces a probability density estimate, which can be difficult to interpret intuitively.
Moreover, it highlights where points are concentrated but does not offer direct insight into their proximity to a central reference (such as the median). This makes it harder to assess how far the errors deviate from the expected range.
Finally, because KDE smooths the data globally, it often under-represents outliers, which may be visually lost in the background despite their importance in evaluating model robustness.

Another way of representing the data could be by using a hexbin plot, as illustrated in Figure~\ref{fig:hexbin_example}. Each hexagonal cell aggregates multiple data points and is colored according to the number of points it contains, with warmer colors indicating higher density regions. This approach effectively addresses the overplotting problem inherent in large datasets and provides a clear view of where most error pairs are concentrated.
However, hexbin plots can be limited for model comparison tasks. The main drawback is that they do not provide any notion of distance or orientation relative to the center of the distribution.

To address this issue, we propose an alternative visualization method that represents points using a colormap based on their proximity to the center of the distribution, which is chosen as the median of the distribution. This approach not only simplifies the interpretation but also provides understanding of the distribution of data points. As a result, this aggregation process tends to smooth the representation and can hide the presence of outliers or extreme deviations that might be critical for evaluating model robustness.
In the visualization shown in Figure~\ref{fig:proximity}, each point is colored based on its distance from the median value. The colormap also reflects the percentile of points, indicating the density of data around the median. Points closer to the median are depicted in warmer colors (e.g., red, orange), whereas those further away appear in cooler colors (e.g., blue). Additionally, a white boundary marks the region where the number of points inside equals the number of points outside this crown, providing a visual clue to quickly identify the core distribution of the data. This helps in understanding the spread and density of points relative to the median.
The visualization of density becomes particularly relevant when one of the distributions contains outliers, namely points that deviate strongly from the central mass due to unusually large or small errors. Traditional representations struggle to reveal these cases, whereas combining density and distance makes them immediately apparent.

% On the dataset used here, most error pairs concentrate very close to zero, despite the extremely wide range of values (from –5000 to 5000). A small number of points lie much farther away, around 10 000 to 15 000, and represent less than 5–10\% of the data. By mapping proximity to the median using a color scale, these rare but critical outliers clearly stand out from the dense central region, even though the raw errors might initially appear reasonable. Without this density-based encoding, such extreme deviations would remain largely hidden.

\subsection{Distances}

The notion of proximity can be calculated using various distances, with the Euclidean distance being the most common. However, this distance primarily highlights differences in scale rather than capturing the true distribution of the data. Unlike Euclidean distance, the Mahalanobis distance accounts for correlations between variables and differences in their scales~\cite{mahalanobis}. It measures the distance between a point and a distribution, allowing for a deeper understanding of the data structure and making it particularly useful for identifying outliers.
% The Mahalanobis distance $D_M$ is defined as: $D_M(x) = \sqrt{(x-\mu)^T \Sigma^{-1}(x-\mu)}$, where $x$ is the vector corresponding to the observation, $\mu$ is the mean vector of the distribution, and $\Sigma^{-1}$ is the inverse of the covariance matrix.

\begin{figure}[htbp]
    \centering
    \includegraphics[width=0.5\columnwidth]{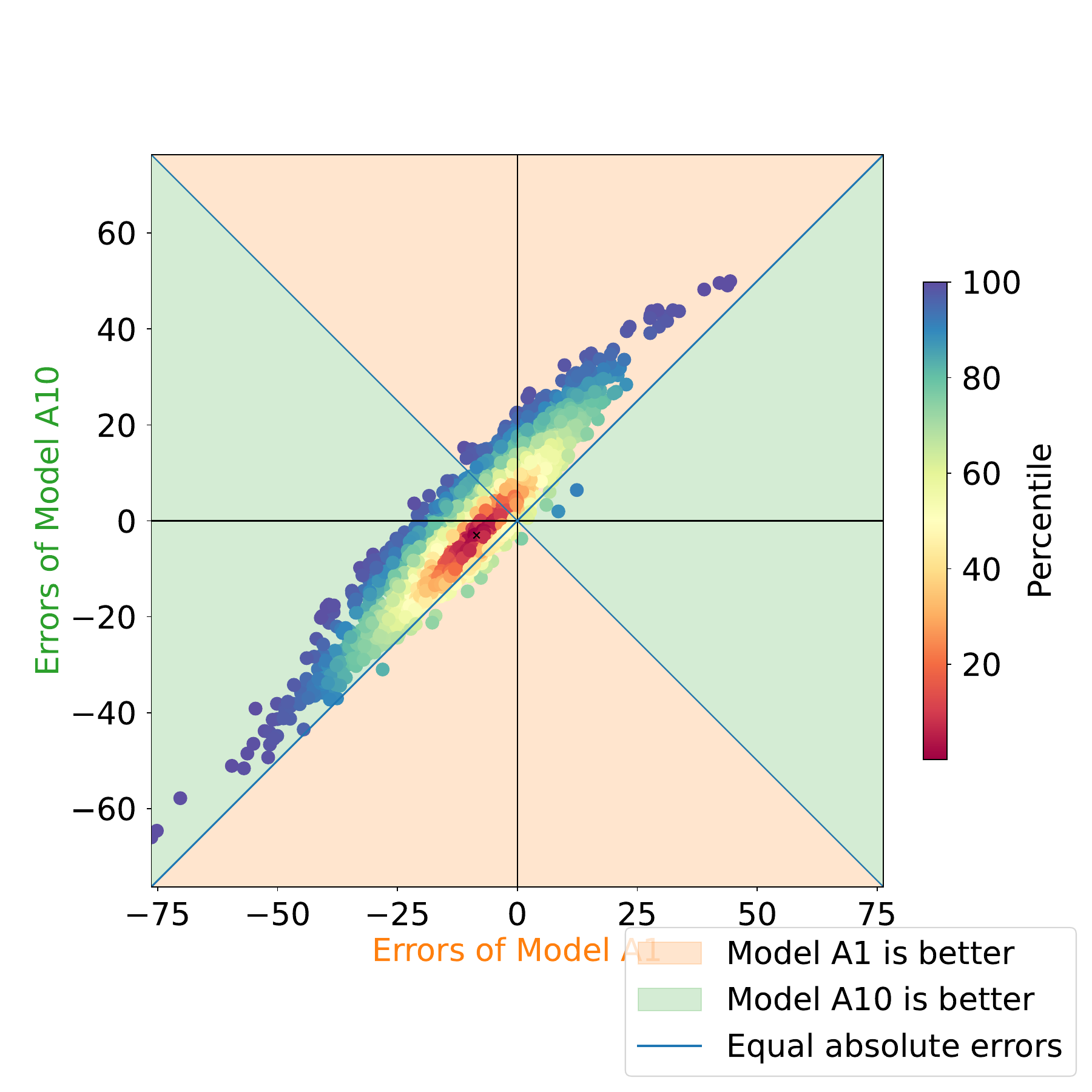}
    \caption{Visualization with Mahalanobis distance for Model A1 and A10} 
    \label{fig:comparison_distance}
\end{figure}

In Figure~\ref{fig:comparison_distance}, the plot illustrates a distribution concentrated near the left diagonal, indicating that the two models share common errors. However, these errors are sparse, ranging from -75 to 50.
When using Euclidean distance to measure proximity to the median, most points appear evenly distributed around a circular region with a radius of approximately 10. In contrast, the Mahalanobis distance reveals a different pattern: the majority of points lie within an elongated ellipse with a narrow semi-minor axis and a wide semi-major axis. This emphasizes the disparity along the diagonal and highlights the variability and correlations in the data that Euclidean distance fails to capture.
By accounting for both the scale and correlation of variables, Mahalanobis distance better reveals such outliers and provides a clearer understanding of the true distribution of the data. Furthermore, in 2D and considering $n$ points, its computational complexity is identical to Euclidean distance, namely $O(n)$.

\section{Case Study}

To demonstrate our visualization methodology and the limitations of traditional metrics, we utilize the AI4I 2020 Predictive Maintenance dataset~\cite{ai4i2020}. This dataset comprises 10,000 synthetic observations of industrial machinery, including categorical product types and continuous physical measurements such as temperature, torque, and rotational speed. The primary regression objective is to estimate the Remaining Useful Life (RUL) of components. In this context, prediction errors carry asymmetric costs: over-estimating RUL is critical as it risks unexpected failure, whereas under-estimating merely triggers premature maintenance.

\subsection{Experimental Setup}

Our experimental setup isolates these asymmetric effects by comparing two neural networks sharing an identical architecture (two hidden layers with 128 and 64 neurons, ReLU activations, dropout p=0.2) and preprocessing steps (80/20 train–test split). The models differ solely in their loss function configuration, utilizing the QUAD-QUAD loss~\cite{rengasamy_asymmetric_2020} to enforce different penalty structures.
The first model uses a low asymmetry parameter ($a=0.2$), strongly penalizing over-estimation—about six times more than under-estimation—thereby encouraging conservative predictions. The second model uses a higher parameter ($a=0.8$), producing a more balanced penalty where over-estimation is only twice as costly, allowing larger predictions.
This controlled setup isolates the effect of asymmetric error costs, since both models share identical data, architecture, and preprocessing (one-hot encoding of Type, removal of non-predictive variables such as UDI and IDs). The dataset is split into 80/20 train–test partitions with a fixed seed, and both models are trained for 20 epochs. Hyperparameter tuning is intentionally limited, as the objective is to illustrate realistic trade-offs practitioners face when selecting between models.

\subsection{Results and Discussion}

\begin{figure}[htbp]
    \centering
    \includegraphics[width=0.5\columnwidth]{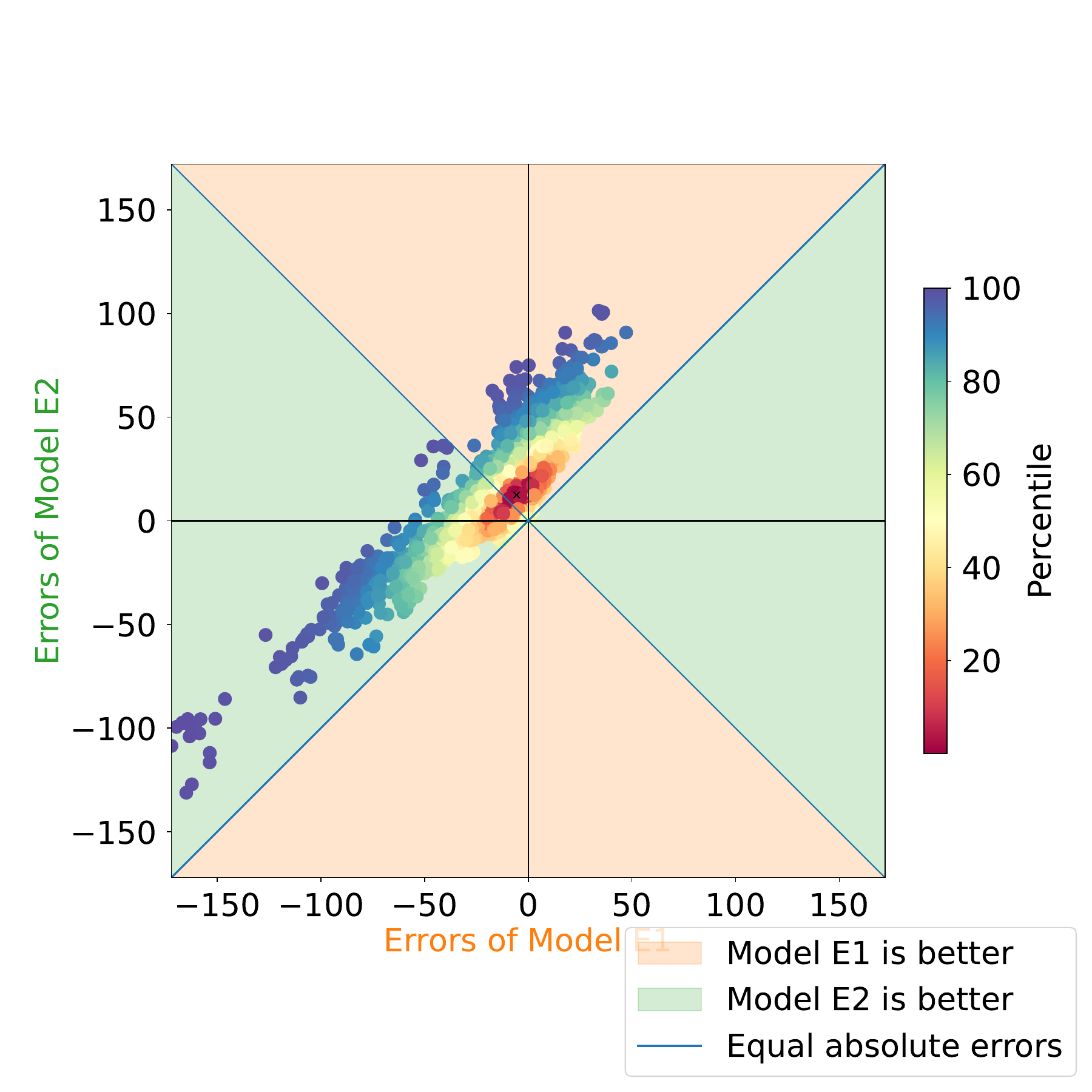}
    \caption{2D Error Space for AI4I2020} 
    \label{fig:ai4i2020}
\end{figure}

The standard performance metrics indicate that Model E1 outperforms Model E2.
Model E1 obtains a lower MAE (20.49 vs. 25.58) and a slightly lower RMSE (32.85 vs. 33.55), reflecting more accurate predictions. Its coefficient of determination is also marginally higher (0.14 vs. 0.10), showing a modest but consistent improvement over Model E2.
However, the 2D Error Space shown in Figure~\ref{fig:ai4i2020} reveals a crucial structural difference between the models that both metrics obscure. The error distribution forms an elongated cloud along the ascending diagonal, indicating that the models are strongly correlated: they struggle on the same individuals. Notably, the majority of the points lie slightly above the identity line $y=x$. This geometric shift signifies that the error of Model E2 is systematically arithmetically larger than that of Model E1.
This behavior aligns with the training setup: Model E1 is conservative, systematically under-estimating to avoid dangerous over-estimations, whereas Model E2 is more optimistic. 
Hence, the visualization of the distribution of paired errors confirms to choose Model E1 if the priority is to minimize unexpected failures.

\section{Conclusion}

Effective techniques for visualizing regression models are essential for their comparative analysis. In this work, we proposed a two-step visualization methodology with three components designed to explore and contrast model performances beyond aggregate metrics like MAE or RMSE.
We first introduced one-dimensional visualizations to assist in selecting relevant models based on their individual error distributions. Then, we proposed a two-dimensional error space that allows for direct comparison between two models by plotting paired errors. This space is enriched with a colormap to represent the distance from the median, offering complementary views on the structure and distribution of errors, including outliers. Finally, we incorporated the Mahalanobis distance to take into account the correlation between error axes to ensure a more robust interpretation of spatial relationships in the comparison zones.
To complement the comprehension provided by these visualizations, we aim to extend this visual toolkit by integrating how model errors evolve across domain or operational conditions by selecting a criterion to visualize model errors. This will further enhance the interpretability of model behavior in dynamic or real-world environments, supporting better-informed model selection and monitoring decisions.

\begin{credits}

\subsubsection{\discintname}
The authors have no competing interests to declare that are
relevant to the content of this article.
\end{credits}

%%
%% Bibliography
\bibliographystyle{splncs04}
\bibliography{references}

\end{document}